\documentclass[lettersize,journal]{IEEEtran}
\usepackage{amsmath,amsfonts}
\usepackage{algorithmic}
\usepackage{algorithm}
\usepackage{array}
\usepackage[caption=false,font=normalsize,labelfont=sf,textfont=sf]{subfig}
\usepackage{textcomp}
\usepackage{stfloats}
\usepackage{url}
\usepackage{verbatim}
\usepackage{graphicx}
\usepackage{cite}
\makeatletter
\def\ps@IEEEtitlepagestyle{%
  \def\@oddhead{}%
  \def\@evenhead{}%
  \def\@oddfoot{}%
  \def\@evenfoot{}%
}
\makeatother

\begin{document}

\title{Simulation of Active Soft Nets for Capture of Space Debris}

\author{Leone Costi and Dario Izzo
\thanks{All authors are part of the Advanced Concepts Team, European Space Agency, Noordwijk, Netherlands, 2201 AZ
        }
}



\maketitle
\pagestyle{empty}

\begin{abstract}

In this work, we propose a mission concept for the removal of orbital debris based on active compliant nets. Conversely from tethered nets that rely solely or mainly on ballistics, we investigate the usage of larger satellites with much higher control authority, able to achieve debris capture in a wide range of approaching conditions. The study is performed through a MuJoCo-based simulator that includes net dynamics, contact between the net and the debris, self-contact of the net, orbital mechanics, and a controller for the thrusters on the four satellites at the corners of the net. We showcase the case of capturing Envisat, whilst investigating different mechanical models and control strategies. Moreover, we do not assume that the net has been previously ballistically thrown toward the target nor that is tethered to a mother spacecraft, and instead we solely rely on the four satellites. The results show that increased compliance results in added maneuverability. Additionally, when paired with a sliding mode controller, soft nets are able to achieve successful capture in $100\%$ of the static cases, whilst also showcasing a higher effective area at contact. When investigating the effect of Envisat's tumbling, we show that we are able to achieve capture in a wide range of approaching conditions, except for when the tumbling axis is perpendicular to the approaching direction.
\end{abstract}


\section{Introduction}
Satellites have moved from specialist assets to largely unseen infrastructure, supporting essential services across communications, navigation, finance, and security as mega‑constellations expand at an unprecedented pace. While Europe’s Galileo \cite{oehler2006galileo} is maturing into a civilian‑controlled navigation backbone, deployments led by Starlink and other large constellation producers have placed thousands of spacecraft in orbit, reshaping expectations of continuous, low‑latency, global coverage \cite{michel2022first}. This abundance has made orbital availability an implicit assumption for an ever‑wider range of terrestrial and spaceborne applications, precisely as its reliability faces mounting strain. The rapid growth in LEO activity is colliding with a deteriorating debris environment \cite{Dhinakaran2025}, where escalating collision risk and routine avoidance are eroding mission assurance today and threatening the long‑term sustainability of key orbital regimes \cite{bigdeli2025mechanics}. Moreover, the monitoring and control of debris is no longer sufficient to ensure safety in LEO \cite{Dhinakaran2025}, and active removal is often advised in order to lower the risk of collisions \cite{liou2011active}. As an example, the ESA e.deorbit mission had the purpose of de-orbiting a single large ESA-owned space debris from the LEO protected zone\cite{biesbroek2017deorbit}. Despite the cancellation of the mission, several studies were conducted on possible active debris removal concepts that could have been utilized. Active debris removal can be achieved with several robotic platforms, ranging from close approach methods such as grippers, harpoons, and nets to contact-free options such as lasers and electro-magnetic fields \cite{Arshad2025, bigdeli2025mechanics}. Both rigid grippers \cite{Dawood2024} and harpoons \cite{Zhao2022} can result in mechanical damage to the debris, potentially breaking the debris itself into multiple pieces, worsening the problem. Soft grippers have been hypothesized as an alternative to rigid ones \cite{ruiz2024thermally}, but the low durability of state-of-the-art elastomeric materials prevents them from being a valid alternative. Contact-free options such as lasers present challenges in efficiency and costs \cite{phipps2012removing}.

In this work, we focus on methods based on soft nets. The dynamics involved have already been investigated in detail in both the deployment phase \cite{botta2017energy} and the capture phase \cite{botta2017contact}, as well as the relative control from a distance post-capture \cite{Field2022, Field2023}. Given the importance of modeling and simulation before deployment \cite{botta2019simulation}, many attempts have been performed at implementing a simulator, from the Bullet physics engine \cite{wormnes2013throw} to Matlab \cite{benvenuto2015dynamics}. These two works have later been merged into a complete simulator that was verified with a parabolic flight experiment \cite{cercos2014validation}. Concurrently, several other simulators have been deployed, with a focus on different phases of the capture process. Some studies are centered on the variable dynamics of the debris \cite{benvenuto2015dynamics, shan2018contact}, whereas others have decided to investigate the robustness of the capture phase \cite{shan2020analysis}. Moreover, \emph{Si et al.} \cite{si2019dynamics} have investigated the effect of the self-contact of the net, often ignored by other studies. Finally, \emph{Botta et al.} \cite{botta2020simulation} presented a software simulation tool for modeling the ballistic capture of space debris through a tether-actuated net system with active closure control. Beyond single-debris capture scenarios, prior studies have explored systems targeting substantially larger objects, including asteroid-scale bodies \cite{zhang2022asteroid}, as well as mission concepts aimed at capturing and removing multiple debris objects within a single operational campaign \cite{boonrath2025mission}. 

Many prior works on tether-net capture of space debris have investigated explicit mechanisms to close the perimeter of the net after contact, with the objective of mechanically securing the target and preventing escape during subsequent towing or de-orbiting phases. Commonly studied solutions rely on tether-actuated closure systems, in which one or more closing lines are routed along the net mouth and reeled in by winches after impact, either through interlaced perimeter threads or multiple attachment points to each side of the net \cite{Benvenuto2016,Botta2019,Sharf2017}. Variants of these concepts include split-mass or thread–ring mechanisms designed to ensure reliable closure regardless of activation timing \cite{Si2021}, as well as separable or impulsively actuated closing devices validated through ground experiments and high-fidelity simulations \cite{mengsheng2025}. Several studies have integrated such closing mechanisms into multibody or finite-element simulators to assess closure dynamics, tension distribution, and containment robustness under different capture scenarios \cite{botta2020simulation}. In contrast to these approaches, the concept proposed in this work does not rely on a mother spacecraft nor on an explicit net-closing mechanism. Instead, capture is achieved through four actively controlled satellites located at the corners of the net, which continuously regulate their relative positions to maintain confinement of the target. The objective is not to mechanically close the net perimeter, but to actively drive and keep the satellites sufficiently close to prevent loss of Envisat, while retaining full control authority to perform subsequent maneuvers, including reconfiguration and de-orbiting.

Although most of the research field relies on fully passive nets, active control of the four corners has been proposed as an alternative method to further improve the success rate of capture. Explored strategies of active control include maximizing the net's area at contact \cite{ru2022capture}, considering multiple closing actions in case of initial failure \cite{zhao2019capture}, sub-optimal contact point \cite{zhao2020capture}, and dynamical assessment of the target position \cite{zhao2022dynamic}. Despite using active control, these studies still rely on a mother-ship to ballistically launch the net toward the target, and the control is solely used to perform small adjustments or to help the closing of the net around the debris after the contact is initiated. Conversely, we propose a mission approach in which the four satellites are much heavier and have more control authority, thus can be used without a close by mother spacecraft or a tether.

In this work, we present a simulator\footnote{\url{https://gitlab.com/}} that can reproduce the physical interaction between the net and the debris, as well as the self-contact of the net, and includes modules both for the closed-loop control strategy and the effect of orbital dynamics. Moreover, we showcase how the simulator can be used by implementing the case of the capture of Envisat \cite{louet1999envisat}. Envisat was a European Space Agency Earth-observing satellite launched in 2002, and after losing contact in 2012, it now remains inactive in orbit around Earth as a large piece of space debris. As one of the biggest debris present in LEO, it represents a particularly challenging milestone for the field of space debris removal \cite{estable2020capturing}.

\section{Material and Methods}

\begin{figure}[htbp]
    \centering
    \vspace{2mm}
    \includegraphics[width=\columnwidth]{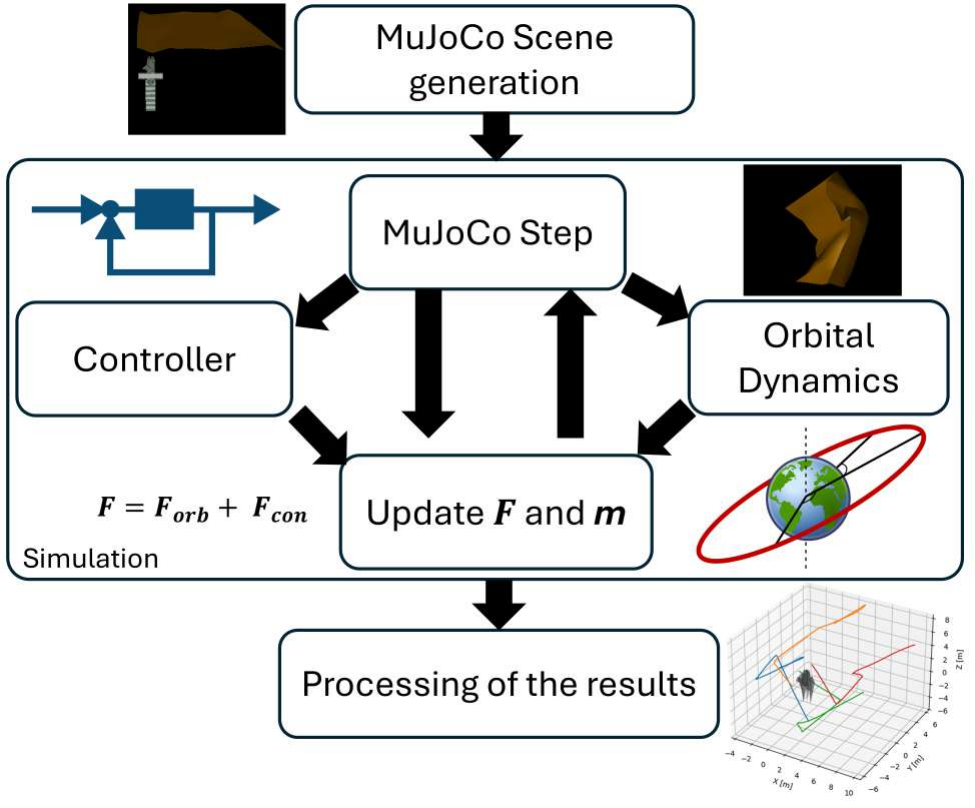}
    \caption{Overview of the simulation tool's workflow. After an initial scene initialization, both the controller and the orbital dynamics modules update the force applied to the massive bodies in the simulation between every step.}
    \label{overview}
\end{figure}

The proposed simulator is based on the open-source physics engine MuJoCo, with added modules to implement orbital mechanics and control strategy. The net is modeled as $N \times M$ point masses connected by material-defining constraints in a grid-like fashion.  Fig. \ref{overview} shows how the simulator merges the contribution of the active controller and the action of orbital dynamics, updating the applied force on all objects in the scene, between the physical simulation steps.
Practically, additional forces $\mathbf F$ are computed and added to the simulation each timestep, and can be divided as follows:
\begin{equation}
\begin{aligned}
\mathbf F &= \mathbf F_{orb} + \mathbf F_{con}\\
\end{aligned}
\end{equation}
where $\mathbf F_{orb}$ is the added force acting on each point mass due to orbital mechanics and $\mathbf F_{con}$ is the thrust applied by the control strategy. Additionally, the mass of the four thrusters is also updated in between simulation steps, in order to account for the mass ejected during propulsion.
\subsection{Physics Simulator}

MuJoCo itself is used to simulate Envisat and the soft net. The specific dynamics used for the connections are implemented with three different methods, as to be able to represent three different continuous materials: Inextensible edges, Shell, and Saint-Venant solid. In Inextensible edges, the distance between connected nodes is fixed, but the nodes can act as pivots, achieving large bending behavior. Shell is similar to the previous case but with a user-defined bending stiffness that limits the local bending radius, achieved with the Shell plug-in. Conversely, Saint-Venant models the net as a Saint-Venant solid thanks to the Elasticity plug-in, based on user-defined values of Young's Modulus, Poisson ratio, and damping. These three different models are used to simulate different mechanical behaviors. Depending on the material used in the real scenario, the net can be simulated as extensible, inextensible, but infinitely flexible, or just flexible. Additional user inputs that can be used to tune the simulation are the 3D object used as debris, its initial velocity, and the relative initial position and velocity of the net.

Contact interactions between the net and Envisat, and self-contact of the net are modeled using MuJoCo’s default constraint-based contact formulation. Each contact is represented by a three-dimensional constraint, consisting of one normal non-penetration constraint and two tangential constraints spanning the contact plane. Normal forces are enforced as unilateral constraints with compliant regularization, while tangential forces implement a velocity-level approximation of Coulomb friction. Contact forces arise as Lagrange multipliers and are solved implicitly within a unified constrained optimization problem at each simulation step, together with joint and actuation constraints, enabling stable resolution of multiple simultaneous and persistent contacts.

\subsection{Orbital Mechanics}
\label{om}

To simulate the effect of orbital mechanics, we implement the Yamanaka-Ankersen (YA) equations, which describe the linearized relative motion in an elliptical reference orbit. The YA formulation generalizes the Clohessy-Wiltshire (CW) model by accounting for the time-varying orbital radius and angular velocity \cite{yamanaka2002new}.

Envisat orbits at an approximate altitude of $790\,km$, corresponding to a mean orbital radius $r_0 = 7{,}171\,km$. The orbit is nearly circular, with an eccentricity on the order of $10^{-3}$. Under these conditions, the YA model reduces to dynamics that are very close to the CW equations.

The YA equations describe the relative dynamics in the Hill (Local-Vertical Local-Horizontal) frame centered on Envisat:

\begin{equation}
\begin{aligned}
\ddot{x} &= 3\frac{\mu}{r^3} x + 2\dot{f}\,\dot{y} - \gamma \dot{x}, \\
\ddot{y} &= -2\dot{f}\,\dot{x} - \gamma \dot{y}, \\
\ddot{z} &= -\frac{\mu}{r^3} z - \gamma \dot{z},
\end{aligned}
\end{equation}
\noindent
where $\mu$ is the Earth's gravitational parameter, $r$ is the instantaneous orbital radius, $\dot{f}$ is the angular velocity associated with the true anomaly $f$, and $\gamma = \dot{r}/{r}$ represents the normalized radial rate induced by the orbital eccentricity.

The variables $x$, $y$, and $z$ denote the relative position of a point mass with respect to Envisat in the radial, along-track, and cross-track directions, respectively. Since the net is simulated using point masses, the YA accelerations are computed for each mass treated as a chaser body. The equivalent orbital force is therefore defined as

\begin{equation}
\mathbf F_{orb} = m [\ddot{x}, \ddot{y}, \ddot{z}]
\end{equation}

where $m$ is the mass of the corresponding point mass.

Given the extremely small eccentricity of the Envisat orbit, the time-varying contributions of the YA model are negligible for the purposes of this study. Consequently, all simulations are performed assuming $f = 0$ at the start of the simulation.

\begin{table}[t]
\centering
\caption{Mass properties of the Envisat capture system}
\label{tab:mass_properties}
\begin{tabular}{|l| c| c|}
\hline

\textbf{Component} & \textbf{Quantity} & \textbf{Mass} \\
\hline
\hline
Envisat (target debris) & 1 & 7821 kg \\
Corner satellites & 4 & 350 kg each \\
Soft net & $10 \times 10$ nodes & 10 kg total \\
\hline
\end{tabular}
\label{mass}
\end{table}

\subsection{Control Strategy}

\begin{figure}[htbp]
    \centering
    \includegraphics[width=\columnwidth]{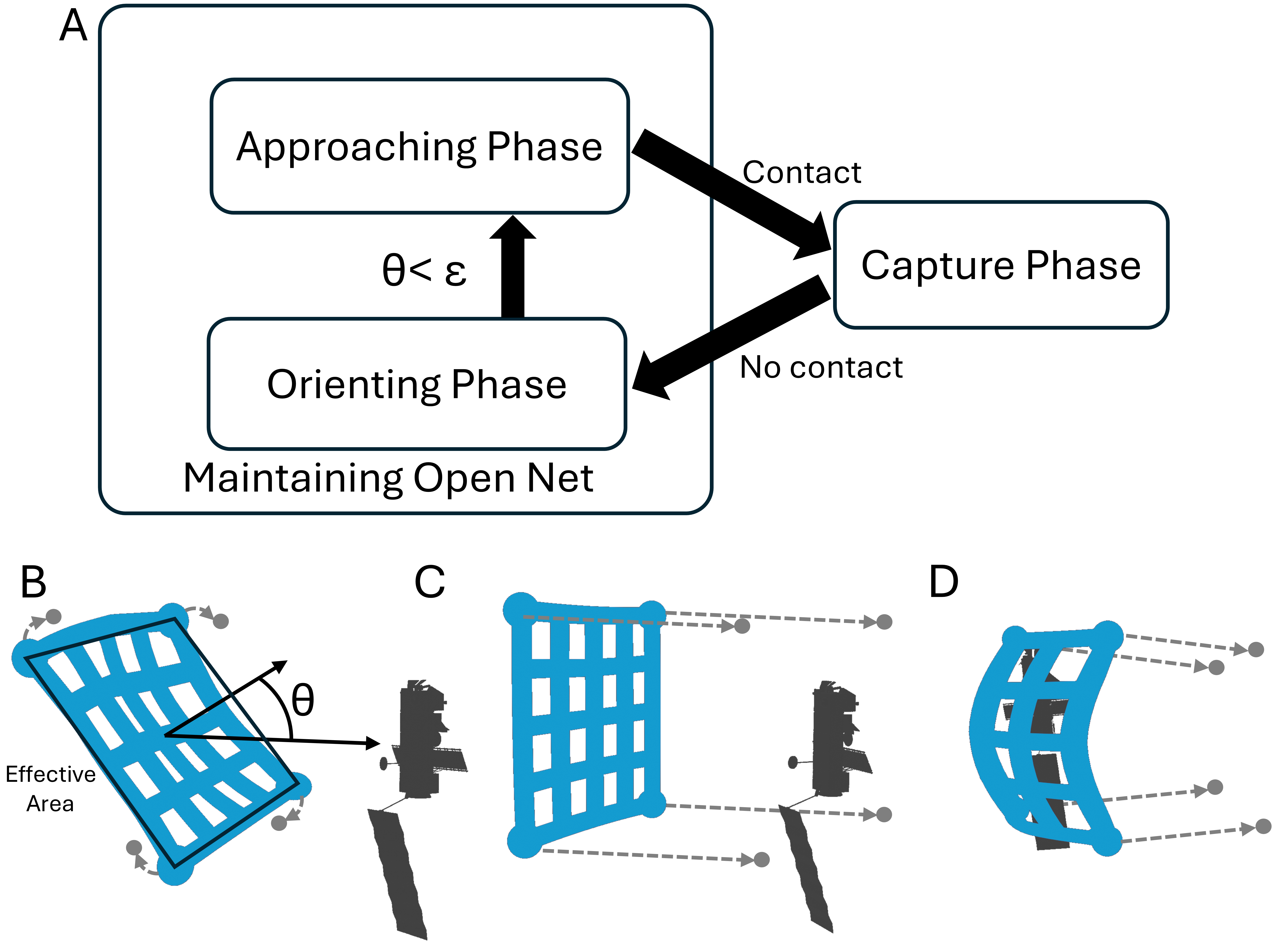}
    \caption{Schematics of the (A) finite-state machine representing the high-level control for the capture of space debris and of the three implemented states: (B) the orienting phase, (C) the approaching phase, and (D) the capture phase.}
    \label{f1}
\end{figure}

The control module receives the position of the net and of the target, Envisat, as inputs and determines the force $\mathbf F_{con}$ used by the control strategy. Whereas in principle every point mass could be controlled, we have limited the control to the four heavier masses placed at the corners of the net, simulating the satellites. This is not only in agreement with the state-of-the-art literature but also represents a realistic scenario for a practical implementation. In the example presented here, the control strategy is divided into high-level and low-level control. The high-level control selects the current phase within the capture and the relative target position of the satellites, and it is implemented as a finite-state machine (see Fig. \ref{f1}).
In a general initial scenario with the net not in contact with debris nor aligned to it at the start, the four satellites at the corners of the net start rotating the net whilst also maintaining an open square configuration to maximize the useful capture area of the net. Such a phase is herein called the orienting phase (see Fig. \ref{f1}B). This is achieved by considering the projection of the satellites onto a plane defined by the first two principal components of their position. Such principal components are extracted using a PCA, and the angle between the normal to the plane defined by those components and the vector from the center of the net to the center of the target is defined as the residual angle $\theta$. In the orienting phase, the target position of the satellites is computed by a rotation of the current position that corresponds to $\theta=0$, whilst maintaining a stretched configuration. When the $\theta$ is below a given threshold $\varepsilon$, which has been set to $36\,deg$ in this work, and the four satellites' projections cover at least $80\%$ of the area of the fully stretched configuration, the finite state machine switches to the approaching phase, where the target position is similar to the previous phase, but with an added component, defined as the vector from the center of the net to the center of the target (see Fig. \ref{f1}C). Lastly, upon contact, the satellites' centroid is determined by their mean position, and they are propelled toward such a centroid to properly envelop the target (see Fig. \ref{f1}D). If the capture fails, the net detects it upon the absence of external contact and starts opening up and aligning with the target, preparing itself for another attempt.

After the high-level control selects the target position of the four corners, the low-level control is implemented either using PID or Slide Mode Control (SMC). The PID controller is implemented as follows:
\begin{equation}
\begin{aligned}
\mathbf{F}_{\text{con}} &= K_p e +  K_i \int_0^T e\;dt + K_d \dot{e}  \\
\end{aligned}
\label{PID}
\end{equation}
where $e=q_{tar}-q$, $q=[x,y,z]$, the index $tar$ refers to the target positions determined by the high-level control, $T$ is the current time, and $ K_p$, $ K_i$, and $ K_d$ are the proportional, integral, and derivative gains, respectively. For this study, these parameters are set to $ K_p = 10^{-2}$, $K_i = 10^{-4}$, and $K_d = 10^{-3}$. Such values have been empirically tuned to achieve a quick response, thus higher maneuverability of the net, whilst maintaining the thrust within a feasible range.

Alternatively, we employ SMC to drive the system toward the target point. SMC is a nonlinear control technique known for its robustness in the presence of modeling uncertainties and external disturbances \cite{utkin2003variable}. This robustness arises from the use of a discontinuous, or rapidly varying, control term that enforces the system's trajectory to ``slide'' along a predetermined manifold in the state space, known as the sliding surface.
In this approach, the control force $\mathbf{F}_{\text{con}}$ is defined as:
\begin{equation}
\begin{aligned}
\mathbf{F}_{\text{con}} &= \mathbf{F}_{\text{eq}} + \mathbf{F}_{\text{sw}}
\end{aligned}
\end{equation}
where $ \mathbf{F}_{\text{eq}}$ and $\mathbf{F}_{\text{sw}}$ are the equivalent control and switching terms, respectively. The former term dictates motion along the sliding manifold toward the target state, while the latter represents a nonlinear switching control that engages whenever the trajectory departs from the manifold, enforcing convergence back onto the surface. The switching term is defined as follows:
\begin{equation}
\begin{aligned}
\begin{cases}
  \mathbf{F}_{\text{sw}} = - K \, \tanh\left(\frac{s}{\sigma}\right) \\
  s =  - (\dot{e} + \lambda e)
\end{cases}
\end{aligned}
\end{equation}
where $s$ is the manifold, or sliding surface, and the parameters $\lambda$, $K$, and $\sigma$ are user-defined control gains that shape the system's convergence behavior and robustness. Specifically, $\lambda$ determines the slope of the sliding surface, $K$ sets the magnitude of the switching control term, and $\sigma$ smooths the discontinuity in the $\tanh$ function to mitigate chattering effects. Note that we maintain the definition of the error as $e = q_{\text{tar}} - q$ for consistency with the PID controller (see Eq. \ref{PID}), rather than $e = q - q_{\text{tar}}$, which is most commonly used in SMC formulations \cite{edwards1998sliding, shtessel2014sliding}. The term $\mathbf{F}_{\text{eq}}$ is then derived by imposing $\dot{s}=0$, leading to:
\begin{equation}
\begin{aligned}
\mathbf{F}_{\text{eq}} &= m \ddot{q} - m \ddot{q}_{tar}  - m \lambda \dot{e}
\end{aligned}
\end{equation}
which finally results in the control law:
\begin{equation}
\begin{aligned}
\mathbf{F}_{\text{con}} &= m \ddot{q} - m \ddot{q}_{tar} - m \lambda \dot{e} - K \, \tanh\left(\frac{\dot{e} + \lambda e}{\sigma}\right)
\end{aligned}
\end{equation}
with positions, velocities, and accelerations both of the target and point masses directly accessible via MuJoCo at each timestep. This formulation enables fast convergence to the desired state while maintaining stability under uncertainties and disturbances, which makes SMC particularly suitable for high-precision or safety-critical tasks. For this study, these parameters are set to $\lambda = 1 \cdot 10^{-2}$, $K = 3 \cdot 10^{-2}$, and $\sigma = 10^{-2}$. In particular, $\lambda$ has been selected identical to $K_p$ for the PID controller, as both terms govern the aggressiveness of error correction: $\lambda$ determines the exponential convergence rate of the tracking error on the sliding surface, in the same way that $K_p$ influences the decay rate of the error in a proportional control scheme.

Regardless of the low-level control method implemented, the maximum force that can be delivered by the thrusters has been limited to $20\,N$. After the thrust $\mathbf{F}_{\text{con}}$ has been computed, the mass of each one of the four satellites is updated to account for the mass that was ejected to achieve such a thrust. Based on Tsiolkovsky's rocket equation \cite{tsiolkovsky1954reactive}, the mass of each satellite is computed as:
\begin{equation}
\begin{aligned}
m= m_0 - \frac{1}{I_{sp}\, g_0} \int_0^T \mathbf{F}_{\text{con}}(t)\, dt
\end{aligned}
\label{russian_guy}
\end{equation}
where $m_0$ is the initial mass of each satellite, $T$ is the current time, $I_{sp}$ is the specific impulse and $g_0=9.80665\,m/s^2$ is the standard acceleration due to gravity at Earth’s surface. In this study, since the maximum thrust is set to $20\,{N}$, the specific impulse is set at $I_{sp} = 250\,\mathrm{s}$ to model a representative chemical propulsion system.

The adoption of chemical propulsion does not inherently introduce structural risk to the net system, as plume interaction constraints are primarily geometric in nature \cite{steagall2024effect, kahn2025electrospray}. The required control forces and torques can be generated without directing exhaust toward the net by appropriately canting the thrusters or distributing thrust among multiple angled units. This design principle is well established in proximity operations such as the gravity tractor, which explicitly rely on inclined thruster orientations to prevent plume impingement on the asteroid surface, while compensating cosine losses through thrust allocation rather than direct in-line firing \cite{lu2005gravitational}.

\subsection{Experimental protocol}
\label{ep}

To evaluate the performance of the simulation tool, we have tested all three constitutive relations considered for the net and both low-level control methods. We have performed two sets of simulations and in all cases the integration step is kept constant at $20\,ms$ and simulations are run until the capture of Envisat. The captured state is reached when the net envelopes Envisat, with contact between Envisat and the net, the four satellites within $8\,m$ of each other, and the difference between the barycenter velocities of the net and Envisat being within $0.1\,m/s$. Such conditions need to be met continuously for at least $200$ simulation steps. For the Shell and Saint-Venant formulations, we have selected a Young's modulus of $10\,kPa$, a Poisson ratio of 0.3, and a damping coefficient of $10^{-2}$.

The first set of simulations is designed to investigate all possible initial orientations of the net relative to Envisat, sampling $200$ starting points uniformly from a sphere of radius $5\,m$ centered on Envisat, while maintaining the starting absolute orientation of the net constant. We do not assume that the net is already traveling toward the target, or that it is correctly oriented. In particular, the net is always initialized with the same absolute orientation and 0 relative velocity with respect to Envisat. Whilst the tested conditions are far from an ideal scenario, we have decided to run this test in order to properly evaluate the performance of the proposed controllers, to demonstrate of the current approach can recover from a wide range of relative positions with respect to the target, and to showcase the autonomy that comes from using four large satellites attached to the net in contrast to previous work that relies on much smaller ones, tethered to a nearby mother spacecraft.

\begin{figure}[htbp]
    \centering
    \includegraphics[width=.6\columnwidth]{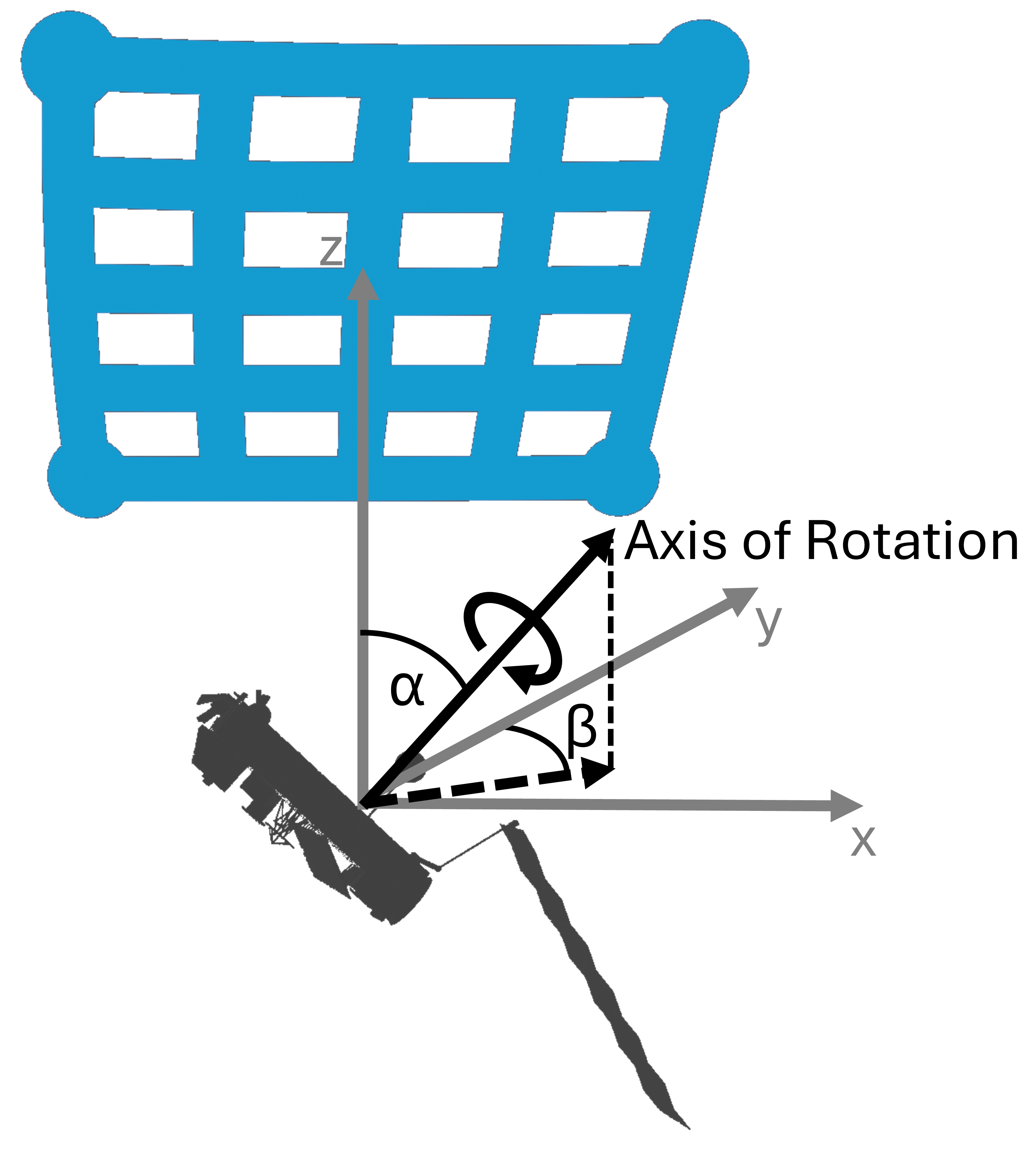}
    \caption{Schematics of the relative positions of Envisat and the net, showcasing the angle $\alpha$ between the axis of rotation and $z$ axis and the angle $\beta$ between the $xy$ plane  projection of the axis of rotation and the $y$ axis.}
    \label{f1_bis}
\end{figure}

The second set of simulations has the main goal of investigating how the controller and the mechanical model of the net perform when Envisat's tumbling is taken into consideration. According to previous studies \cite{pittet2018spin, sagnieres2019long}, Envisat tumbles mainly on its major axis of inertia, and reached a maximum rotational speed of around $3\;deg/s$ in 2013, decreased to $1.6\;deg/s$ in 2016, and slowly increased back to $1.7\;deg/s$ in 2021. In this simulation set, we start with the net $5\;m$ above Envisat, facing it. Envisat has a relative orientation with respect to the net, characterized by $\alpha$, the angle of its highest axis of inertia with respect to the vertical $z$ axis, and $\beta$, the angle in the $xy$ plane, as shown in Fig. \ref{f1_bis}.

\section{Results}
All simulations were performed using a $10\times 10$ point mass grid and an integration step of $20\;ms$. Under these conditions, the simulator runs at frequencies exceeding $50\;Hz$ on a single CPU core, achieving near-real-time execution, a significant improvement over the runtimes reported in previous studies \cite{botta2020simulation}. This computational efficiency is a key enabler for sample-intensive control strategies such as reinforcement learning, where the cost of running high-fidelity simulators has been identified as a critical bottleneck \cite{liu2025surrogate}.

\begin{figure*}[htbp]
    \centering
    \vspace{2mm}
    \includegraphics[width=.85\textwidth]{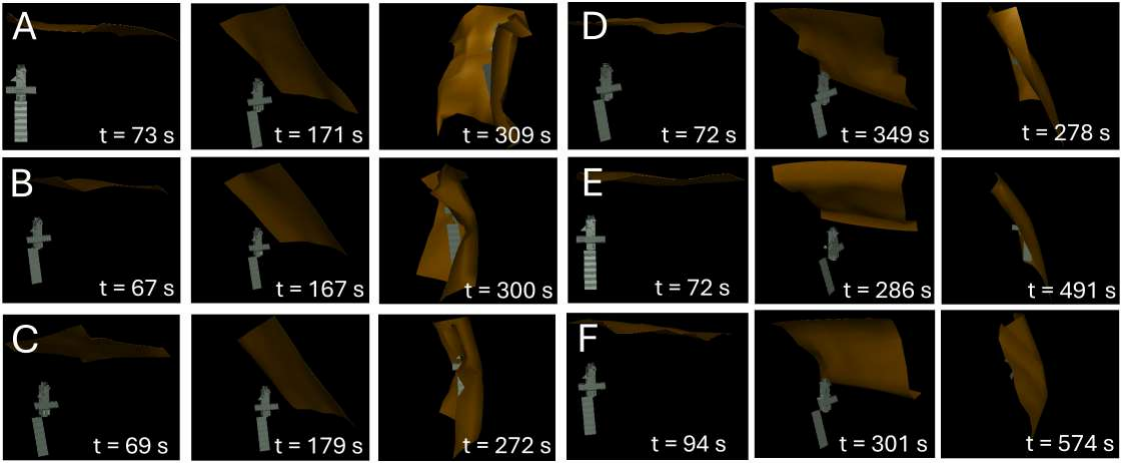}
    \caption{The three phases of the capture of Envisat as a function of the net mechanical model and the controller used: (A) Inextensible edges and PID, (B) Shell and PID, (C) Saint-Venant solid and PID, (D) Inextensible edges and SMC, (E), Shell and SMC (F) Saint-Venant solid and SMC. The starting position of all cases is the same, and the starting relative velocity between the net and Envisat is 0.}
    \label{f2}
\end{figure*}

The capture dynamics vary considerably across mechanical models and control strategies, as illustrated in Fig. \ref{f2}, which shows the three characteristic phases of Envisat capture for each tested configuration. Regarding the control strategy, the SMC controller maintains the desired net opening area throughout the approach, at the cost of a slower initial reorientation: the net undergoes sustained oscillations as the controller enforces the shape constraint, delaying the onset of the translational acceleration toward the target. The PID controller, by contrast, drives the corner satellites more aggressively, producing a faster but less regulated reorientation, with appreciable shape distortion during the capture phase. Regarding the mechanical model, the Shell formulation produces visibly tighter local curvatures compared to the other models, consistent with the additional bending stiffness introduced by this constitutive relation. The Saint-Venant and inextensible edge models exhibit flatter net geometries, reflecting the absence or reduced influence of bending resistance.

\begin{figure}[htbp]
    \centering
    \includegraphics[width=\columnwidth]{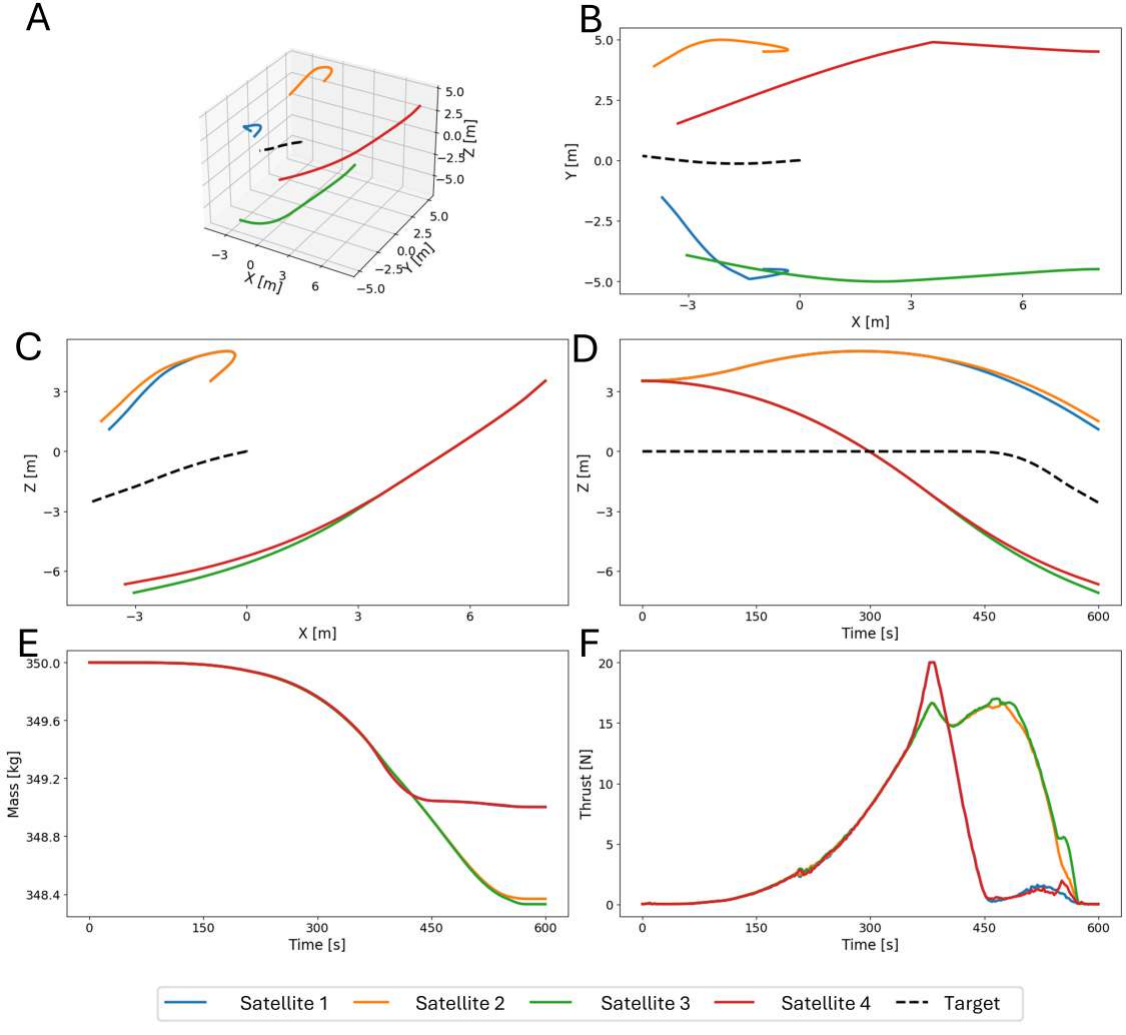}
    \caption{Trajectories of the four satellites with respect to the target: (A) isometric view, projection on the (B) $xy$ plane and (C) $xz$ plane, and (D) $z$ position as a function of time. Data about the thrusters include: (E) mass of the satellites at the corners of the net and (F) thrust generated by each satellite as a function of time.}
    \label{f3}
\end{figure}

For a more in-depth analysis, Fig. \ref{f3} showcases the trajectory taken by the four corners during capture under the Saint-Venant formulation with an SMC controller  over a $500\;s$ simulation.
Figs. \ref{f3}A–C show the three-dimensional trajectories of the target and the net corner satellites, while Fig. \ref{f3}D shows the time history of the $z$ position. The simulation is initialized with zero relative velocity between the net and Envisat; the net subsequently reorients and translates toward the target until capture is achieved. Capture is confirmed by the convergence of the target and net trajectories and by the applied thrust reaching zero, indicating that no further corrective action is required. As reported in Fig. \ref{f3}E, the propellant consumed per satellite is approximately $0.5\;kg$, corresponding to a total of $2\;kg$ for the four-satellite system. This is well within the propellant margin reserved for the capture phase, leaving sufficient budget for the subsequent de-orbiting maneuver.

\begin{figure}[htbp]
    \centering
    \includegraphics[width=\columnwidth]{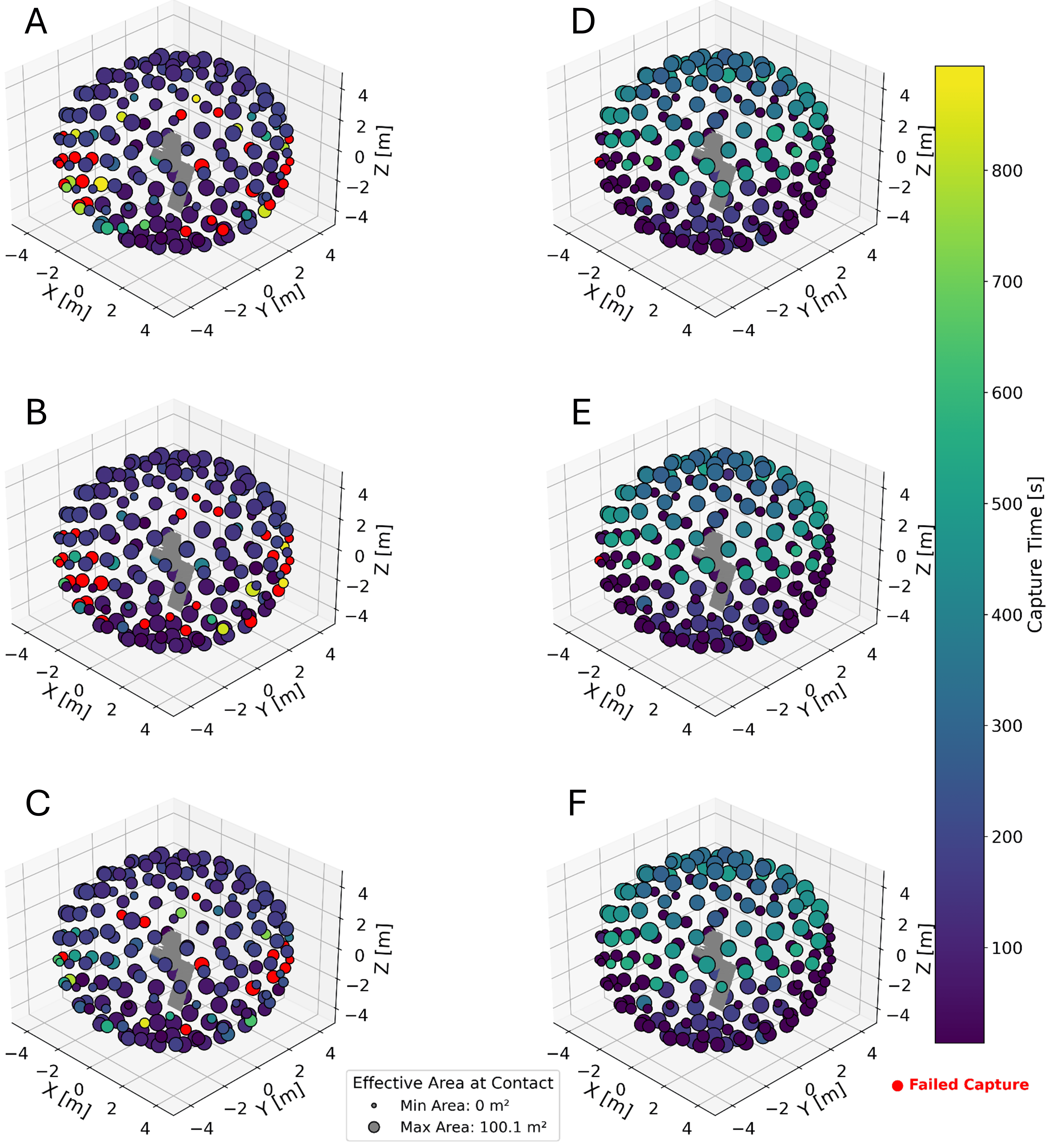}
    \caption{Time needed to achieve capture as a function of the starting positions of the for each control modality and the model used to characterize the net: (A) Inextensible edges and PID, (B) Shell and PID, (C) Saint-Venant solid and PID, (D) Inextensible edges and SMC, (E), Shell and SMC (F) Saint-Venant solid and SMC. The size of the marker is proportional to the effective area of the net upon contact with the target. The effective area is computed by the area between the projections of the four corners onto a plane perpendicular to the vector between the center of the net and the target. Red markers indicate failed capture.}
    \label{dots}
\end{figure}

However, metrics such as the time needed to achieve a capture or the success rate heavily depend on the relative start position between Envisat and the net. Fig. \ref{dots} shows all the tested cases when investigating a wide range of starting conditions, divided by type of controller and model used to represent the net. Additionally, Table \ref{tab:capture_percentage} summarizes the capture percentage, the fraction of all tested starting positions that resulted in a capture within $600\,s$.
\begin{table}[h]
\centering
\begin{tabular}{|l||c|c|c|}
\hline
        & Inextensible edges & Shell & Saint-Venant\\
        \hline
        \hline
PID     &      $88.5\%$             &     $86.5\%$   &             $93.5\%$        \\
\hline
SMC     &          $99.5\%$          &     $99.5\%$   &            $100.0\%$         \\
\hline
\end{tabular}
\vspace{6pt} 
\caption{Capture percentage as a function of the control modality and the model used to characterize the net.}
\label{tab:capture_percentage}
\end{table}

In general, PID performs worse than SMC across all net models. This can be explained by the greater resilience to nonlinearities that SMC shows, thanks to the switching term. This comprehensive view confirms what has been observed for a specific case in Fig. \ref{f2}: SMC is much slower than PID, especially when approaching Envisat from above and from the sides. This difference is less visible when the net starts below Envisat. This is probably due to the presence of Envisat's solar panel. Moreover, SMC also achieves a higher effective area of the net upon contact, indicated by the marker size in Fig. \ref{dots}, potentially achieving an overall higher success rate. For both controllers, the results indicate that more compliance leads to a higher success rate, with the Saint-Venant formulation reaching up to $100\%$ accuracy. Across all successful captures, the average fuel mass needed for the maneuver is $0.4874\;kg$, summing up the total mass expelled by all four satellites, with a maximum value of $9.232\;kg$. Such values are much lower than the $683.7\;kg$ needed for the de-orbiting (see Section \ref{om}), and well within our initially estimated margin.

\begin{figure}[htbp]
    \centering
    \includegraphics[width=\columnwidth]{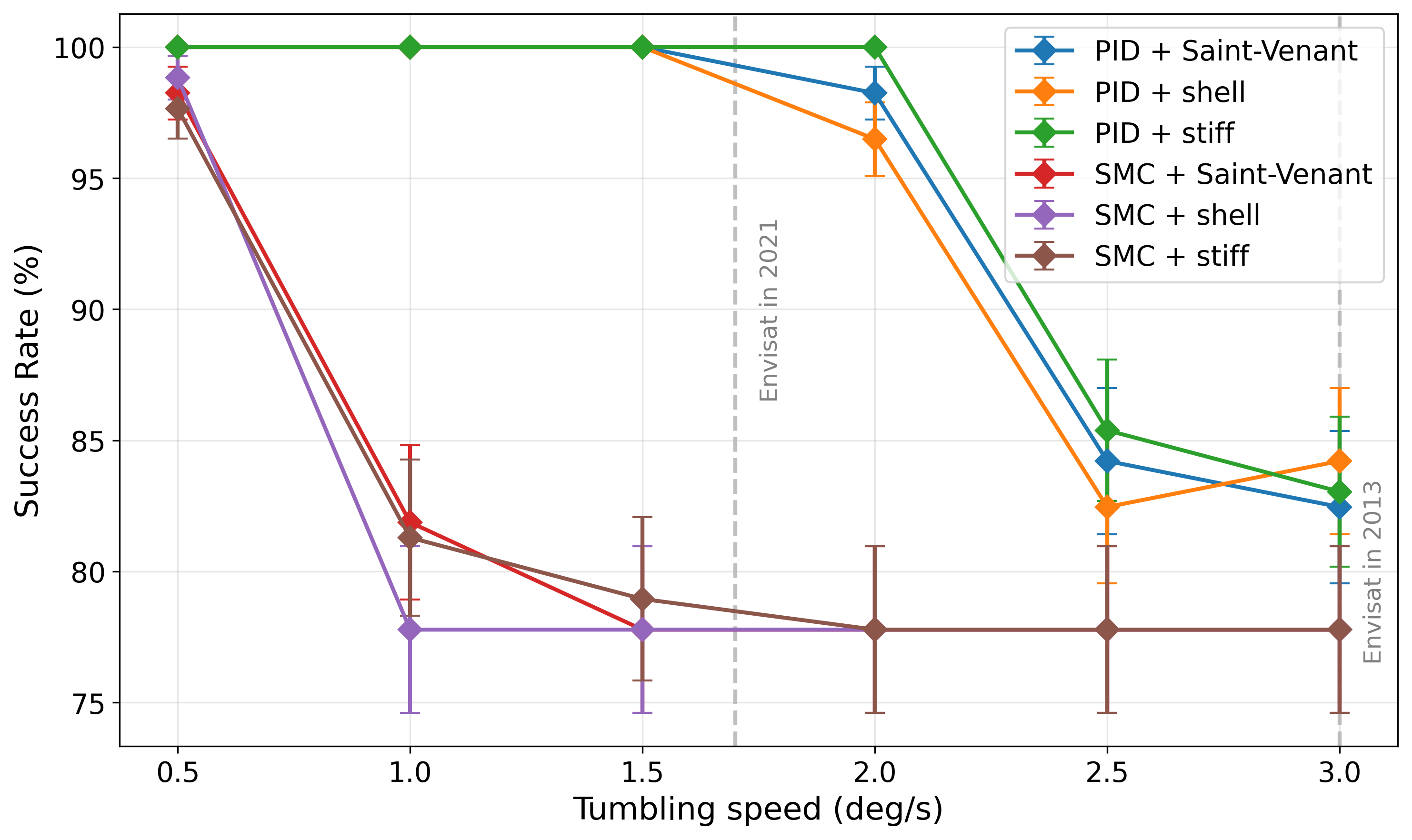}
    \caption{Success rate of capturing Envisat as a function of Envisat's tumbling speed, for the different combinations of controller and mechanical modeling of the net. For reference, the measured tumbling speeds of Envisat in 2013 and 2021 are also reported.}
    \label{s_s}
\end{figure}

Although our results showcase the high level of control authority of the proposed controllers both in the reorienting and propulsion of the net, in a realistic scenario Envisat would be tumbling with respect to the net. Envisat rotates around its major axis of inertia, and such a rotation represents a challenge for the successful capture of the debris. Fig. \ref{s_s} characterized the performance of the different combinations of controller and mechanical model of the net as the tumbling speed of the debris increases.

The data show that the PID controller maintains a much higher success rate with respect to SMC for all scenarios tested. Moreover, the expected drop in performance for high tumbling speeds is not only lower in magnitude, but also significantly delayed, as it appears only for tumbling speeds above $2\;deg/s$. Conversely, SMC implementations, regardless of the mechanical model of the net, start exhibiting a large performance drop as early as $1\;deg/s$.

\begin{figure}[htbp]
    \centering
    \includegraphics[width=\columnwidth]{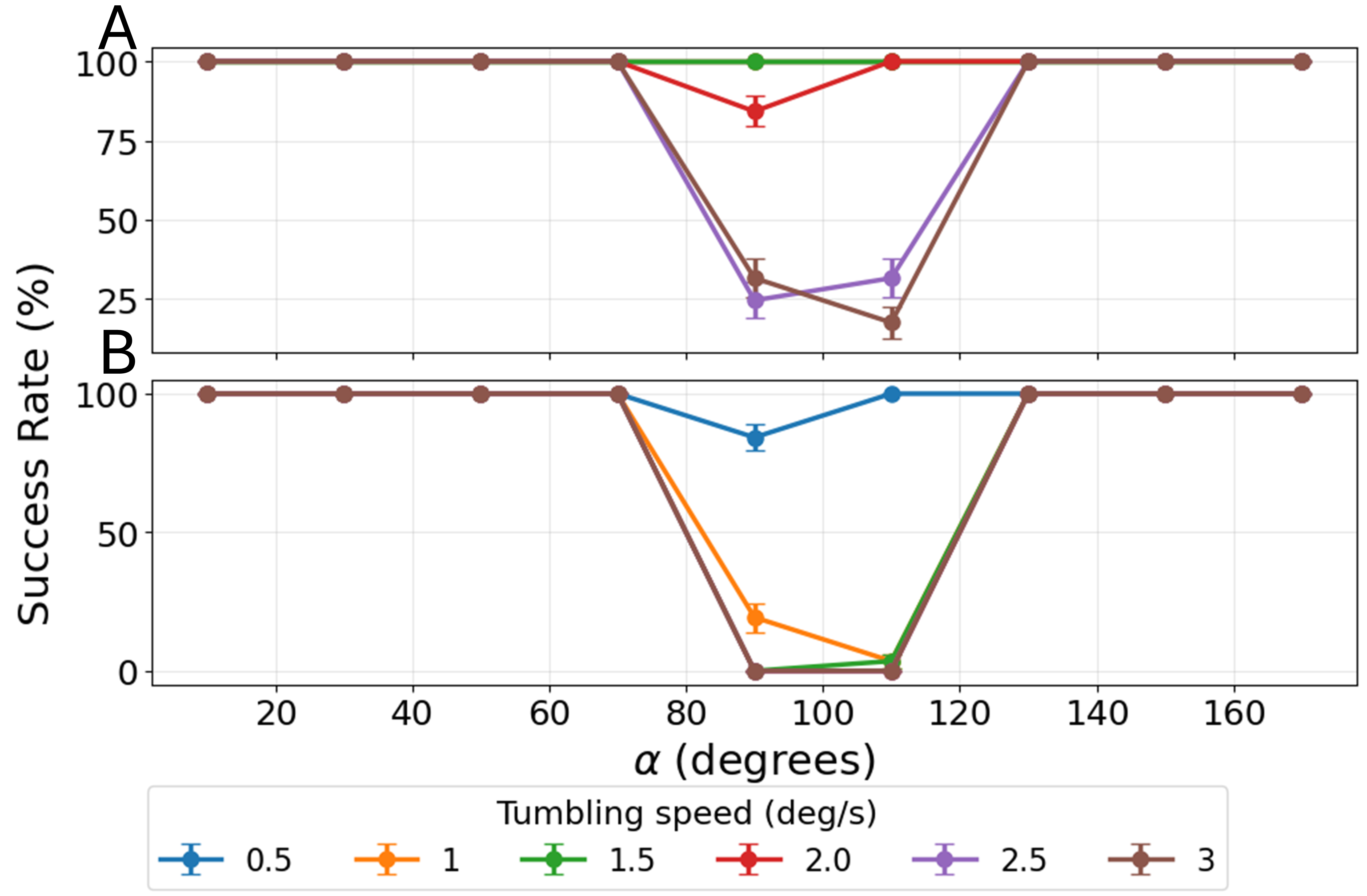}
    \caption{Success rate of capturing Envisat as a function of the angle $\alpha$ between the major axis of inertia of Envisat and the vector pointing from Envisat to the center of the net, for increasing values of tumbling speed and different controllers: (A) PID and (B) SMC.}
    \label{s_a}
\end{figure}

However, the success of the capture largely depends on the approaching direction of the soft net. As described in Section \ref{ep}, we define $\alpha$ as the angle between Envisat's rotational axis and the vector that defines the approaching direction of the net. 
For a wide range of $\alpha$, PID and SMC are physically equivalent, as both reach a success rate of $100\;\%$. The difference in performance only appears around $\alpha=90\deg$, in which the SMC systematically fails the capture and PID is able to complete it only for roughly $25\%$ of the tested values of $\beta$. As a result, the two tested controllers' results are much closer in performance than what was initially suggested by Fig. \ref{s_s}, as they both achieve perfect accuracy for $67\;\%$ of tested cases. Moreover, the angle $\alpha$ is something that can be controlled during the approach. Practically speaking, prior maneuvers can be executed in order to avoid an approach with a value of $\alpha$ near $90\deg$, thus ensuring a successful capture of the target.
Fig. \ref{s_a}A and \ref{s_a}B show the success rate for all mechanical models in the case of PID and SMC controllers, respectively.

\section{Conclusion}

This work presents a simulation framework for the study of debris capture using highly compliant soft active nets, demonstrated through the simulated capture of Envisat. Three constitutive models for the net's mechanical behavior and two control strategies are evaluated across a wide range of initial relative positions between the net and the target. Moreover, since it is implemented in MuJoCo, the proposed simulator is computationally efficient, making it well-suited for large-scale parametric studies and for the future integration of computation-intensive control and learning-based approaches, which are often impractical with current implemented simulators.

The results indicate that increased compliance generally improves capture success rate, as more compliant nets conform more readily to the irregular geometry of Envisat, improving contact and reducing the likelihood of escape. Regarding control strategy, SMC outperforms PID in terms of maneuverability, achieving a higher success rate and a larger effective capture area, at the cost of longer capture times. This advantage is attributed to the greater robustness of SMC to the nonlinearities of the problem. However, SMC degrades more severely than PID at high tumbling rates, particularly when Envisat's rotational axis approaches perpendicularity with the incoming direction, a configuration under which both controllers struggle to achieve reliable capture. In practice, this constraint is not fundamental: prior to the close-proximity approach, the chaser satellites can maneuver to modify the angle between the net's approach direction and the target's spin axis, restoring favorable capture geometry.

Nevertheless, whereas in principle Envisat represents one of the hardest debris to de-orbit, given its size and weight, further studies on different types of debris are needed to investigate how both the mechanical properties of the net and the implemented controller can be improved based on the size, mass, and geometry of the target, as well as its orbit. Moreover, both implemented controllers' performance is highly dependent on the parameters. As an example, when changing the starting distance from $5\,m$ to $10\,m$ without changing the parameters, the success rate drops to $93.5\%$. Hence, we endorse further studies on the characterization of how the controller's parameters can be optimized as a function of both the target and the mechanical model of the net.

In summary, we have proposed a simulator that can be used to investigate soft active nets for space debris capture, and have showcased the case of Envisat as a demonstration, achieving $100\%$ capture success when starting the simulations $5\,m$ from it. We hope that this work will pave the way for further technical development of active soft nets as a solution for space debris removal and orbital robotics.



\section*{Author Disclosure Statement}
The authors declare no conflicts of interest.


 
\bibliographystyle{IEEEtran}
\bibliography{references}

\vfill

\end{document}